\documentclass[10pt,twocolumn,letterpaper]{article}
\usepackage{cvpr}
\usepackage{times}
\usepackage{epsfig}
\usepackage{graphicx}
\usepackage{amsmath}
\usepackage{amssymb}
\usepackage{MYdefs}
\usepackage{gendefs}
\usepackage{listings}
\graphicspath{ {./Images/} }
\usepackage[pagebackref=true,breaklinks=true,letterpaper=true,colorlinks,bookmarks=false]{hyperref}

\cvprfinalcopy 

\ifcvprfinal\fi
\begin{document}
%

%
\title{Complex Events Recognition under Uncertainty in a Sensor Network}

\author{Atul Kanaujia\\
ObjectVideo, Inc.\\
{\small atul.kanaujia@gmail.com}
\and
Tae Eun Choe\\
ObjectVideo, Inc.\\
{\small tchoe@objectvideo.com}
\and
Hongli Deng\\
ObjectVideo, Inc.\\
{\small hdeng@objectvideo.com}
}
\maketitle
\thispagestyle{empty}
\begin{abstract}
Automated extraction of semantic information from a network of sensors for cognitive analysis and human-like reasoning is a desired capability in future ground surveillance systems. We tackle the problem of complex decision making under uncertainty in network information environment, where lack of effective visual processing tools, incomplete domain knowledge frequently cause uncertainty in the visual primitives, leading to sub-optimal decisions. While state-of-the-art vision techniques exist in detecting visual entities (humans, vehicles and scene elements) in an image, a missing functionality is the ability to merge the information to reveal meaningful information for high level inference. In this work, we develop a probabilistic first order predicate logic(FOPL) based reasoning system for recognizing complex events in synchronized stream of videos, acquired from sensors with non-overlapping fields of view. We adopt Markov Logic Network(MLN) as a tool to model uncertainty in observations, and fuse information extracted from heterogeneous data in a probabilistically consistent way. MLN overcomes strong dependence on pure empirical learning by incorporating domain knowledge, in the form of user-defined rules and confidences associated with them. This work demonstrates that the MLN based decision control system can be made scalable to model statistical relations between a variety of entities and over long video sequences. Experiments with real-world data, under a variety of settings, illustrate the mathematical soundness and wide-ranging applicability of our approach. 
\end{abstract}
\section{Introduction}
High-level cognitive reasoning for making decisions entails fusing information in the form of symbolic observations, domain knowledge of various real-world entities and their attributes, and interactions between them. Complex events are difficult to define, primarily due to a variety of ways in which different parts of the event can be observed. Real world event inference therefore requires efficient representation of the complex interplay between the constituent entities and events, while taking into account uncertainty and ambiguity of the observations. Decision making is a complex task that involves analyzing data (of different level of abstraction) from disparate sources and with different levels of certainty, merging the information by weighing in on some data source more than other, and arriving at a conclusion by exploring all possible alternatives. Lack of effective visual processing tools, incomplete domain knowledge, lack of uniformity and constancy in the data, and faulty sensors are some of sources of uncertainty in the data. For example, target appearance frequently changes over time and across different sensors, data representations may not be compatible due to difference in the characteristics, levels of granularity and semantics encoded in data. 

In this work, we adopt Markov Logic Networks (MLN) \cite{DR07}, a generic framework for overcoming the huge semantic gap between the low-level visual processing of raw data obtained from disparate sensors and the desired high-level symbolic information for making decisions on complex events occurring in a scene. MLN provides mathematically sound techniques for representing and fusing the data at multiple levels of abstraction, and across multiple modalities to perform complex task of decision making. MLN uses probabilistic first order predicate logic (FOPL) for representing the decomposition of real world events into visual concepts, interactions among the real world entities and contextual relations between visual entities and the scene elements. It should be noted that while the first order logic formulas may be typically true in the real world, they are not always true. In most domains it is very difficult to come up with non-trivial formulas that are always true, and such formulas capture only a fraction of the relevant knowledge. Despite its expressiveness, pure first-order logic has limited applicability to practical problems of reasoning. Therefore, in MLN framework, complex events and object assertions are defined by hard and soft rules. Each formula has an associated weight that reflects how strong a constraint is. The higher the weight, the greater the difference in probability between a world that satisfies the formula and one that does not, provided that other variables stay equal. In general, a rule for detecting a complex action entails all of its parts, and each part provides (soft) evidence for the complex action. Therefore, even if some parts of a complex action are not seen, it is still possible to detect the complex event using the MLN inference. 
\begin{figure*}[t]
\begin{center}
\includegraphics[width=0.58\linewidth]{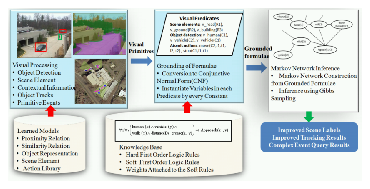}
\includegraphics[width=0.35\linewidth]{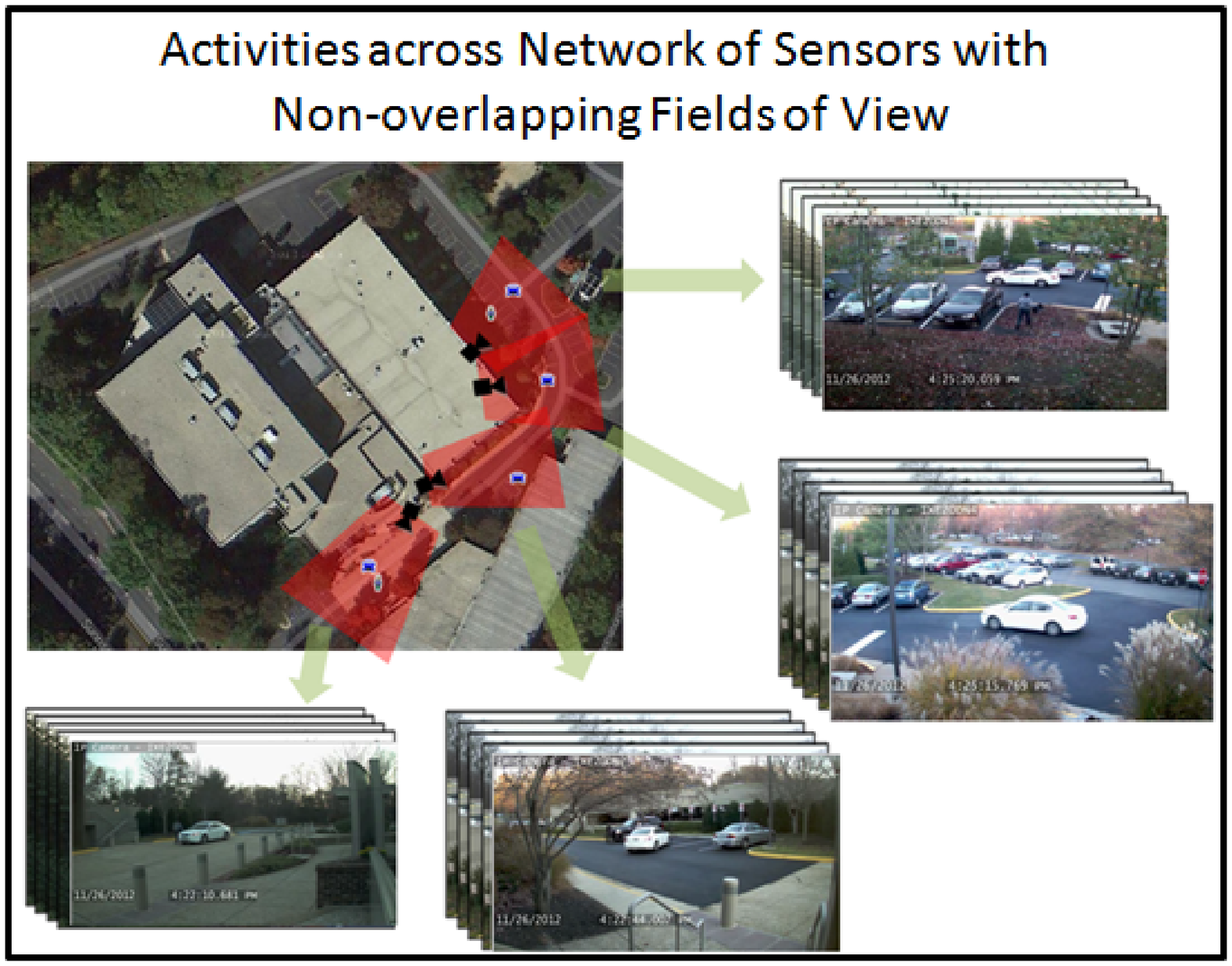}
\end{center}
\caption{{\it(Left)}Overview of the Markov Logic Network(MLN) based decision system for complex event modeling and recognition from synchronize  streams of image sequence, acquired from a network of sensors with non-overlapping fields of view, as shown in the figure on the {\it(Right)}}
\label{fig:MLNProcessing}
\end{figure*}

\noindent\textbf{Related Work:} There have been numerous frameworks for visual event representation and recognition. The frameworks can be broadly divided into declarative approaches \cite{RT00, NZH03} and probabilistic approaches \cite{YOI92}. In declarative approaches events are represented with declarative templates. Events are typically organized in a hierarchy, starting with primitive events at the bottom and composite events on top. The recognition of a composite event proceeds in a bottom-up manner.  These approaches have several drawbacks: (a) A miss or false detection of a primitive event, which occurs frequently in computer vision, not just in crowded or poorly illuminated conditions, often leads to irrecoverable failures in composite event recognition; (b) Uncertainty is often not modeled because of which these methods are generally not robust to typical errors in image analysis. Approaches that are based on probabilistic grammars for event recognition such as \cite{BI98} typically use simple rules. They do not allow existential quantifiers, which are needed for dealing with missing observations. It is also difficult to express domain constraints such as {\it a car can only be driven by one person} using generative grammars. Furthermore, methods to perform probabilistic propagation are better understood for graphical models than for probabilistic grammars. Traditionally Hidden Markov Models(HMM) \cite{BOP97}, Propagation Nets (P-Net)\cite{SBE06} and other forms of Dynamic Bayesian Networks(DBN)\cite{NZH03}\cite{MM07} had been widely applied to event recognition. Being restrictive in terms of number of actors and types of activities that can be modeled due to fixed structure of the model, they require large annotated examples for training. Among rule based activity modeling techniques is the probabilistic method based on multi-agent belief network for complex action detection by Intille and Bobick\cite{IB99}. The method dynamically generates belief network for recognizing complex action using the pre-specified structure that represents temporal relationships between the actions of interacting agents. More recent research has focused on stochastic grammars based event recognition such as \cite{BI98,GSSD09,RA06}. Gupta et al.\cite{GSSD09} developed a storyline model using probabilistic grammars to dynamically infer relations between component actions and also learns visual appearance models for each actions using the weakly labeled video data. Ryoo and Aggarwal \cite{RA06} modeled composite actions and interactions between agents using non-probabilistic Context Free Grammar(CFG). 
Sridhar et. al\cite{SCH10} developed an unsupervised method to identify component events of a complex activity by modeling interactions between subsets of tracks of entities as a relational graph that captured qualitative spatio-temporal relationships between these agents. MLN generalizes over these probabilistic models and offers several advantages over other rule-based activity recognition methods\cite{TCYZ06}. MLN allows ability to write more flexible rules with existential quantifiers over sets of entities, and therefore allows greater expressive power of the domain knowledge compared to other probabilistic rule based methods such as attribute grammars or dynamic Bayesian networks \cite{TD08}. Also methods to perform probabilistic inference are better understood for graphical models used in MLN than for probabilistic grammars. In the past MLN has been applied in the context of scene understanding and activity inference \cite{TD08,SK10,KYD10,MD11}. Tran and Davis\cite{TD08} developed a visual event modeling framework based MLN that addressed a wide range of uncertainties due to detection, missing observations, inaccurate logic rules and identity maintenance. Later works \cite{SK10,KYD10,MD11} further developed the MLN based systems to infer multi-agent activities, use domain knowledge to improve scene interpretation and incorporated Allen's interval logic to improve scalabilty of MLN inference.
  
\noindent\textbf{Contributions:} Our work most closely resembles earlier works \cite{TD08,KYD10} and significantly contributes towards overall understanding and application of MLN in the following way: (a) Unlike past approaches that used hard evidences, our framework models uncertainty at multiple levels of inference, and propagates it bottom-up for more accurate high-level decision making ; (b) We scale MLN inference to more complex activities involving network of visual sensors and increased uncertainty due to inaccurate target associations across sensors ; (c) Apply rule weights learning for fusing information acquired from multiple sensors (target track association) and ; (d) Enhance visual concept extraction techniques using distance metric learning.
\begin{figure}[h]
\begin{center}
\includegraphics[width=0.99\linewidth]{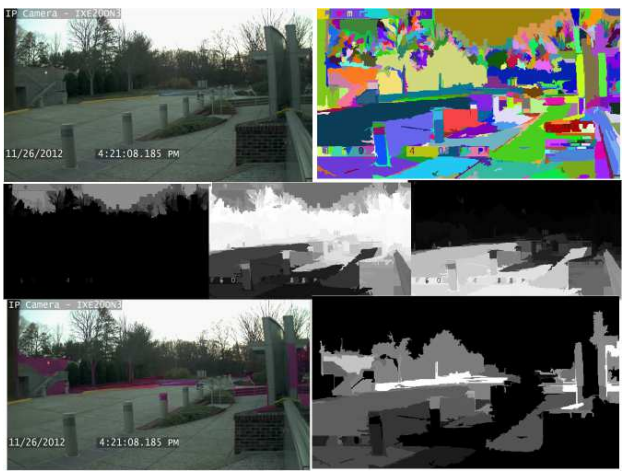}
\end{center}
\caption{Functional scene element labeling: {\it(Top)} Original input image and segmentation results using \cite{FH04}; {\it(Middle)} probabilistic map of segments classified into $\cC =$$\{SKY, VERTICAL, HORIZONTAL\}$ categories ; and {\it(Bottom)} probabilistic map of segments classified as building entry or exit regions, and the corresponding thresholded image segments}
\label{fig:BuildingEntryExit}
\end{figure}
    
\section{System Overview} 
\label{sec:SysOverview}
\Fig{fig:MLNProcessing}{\it(Left)} shows various stages involved in MLN inference. We apply MLN to detect complex activities in a multi-sensor data acquisition and processing scenario as shown in \fig{fig:MLNProcessing}{\it(Right)}. Following components constitute an MLN based decision system. 

\noindent\textbf{Visual Processing:} These modules process videos and extract visual concepts in the form of constants, that denote space-time locations of the entities detected in the scene, scene elements, entity class and primitive events directly inferred from the visual tracks of the entities. The constants are used to ground(instantiate) the variables in the FOPL formulae of MLN. Our visual processing algorithms are composed of detection, tracking and classification of human and vehicle targets, and attributes extraction such as carrying a bag or not. Targets are localized in the scene using background subtraction and tracked in 2D image sequence using Kalman filtering. Targets are classified to human/vehicle based on their aspect ratio. Vehicles are further classified into Sedans, SUVs and mini vans using 3D vehicle fitting\cite{CRTH11} The Atomic events about target dynamics (moving or stationary) are generated from the target tracks. For each event we generate constants for the time interval and pixel location of the target in 2D image (or the location on the map if homography is available). We learn discriminative deformable part-based\cite{FMR08} classifiers to compute a probability scores for whether a human target is carrying a bag. The classification score is fused across the track by taking average of top K confident scores(based on absolute values) and is calibrated to a probability score using logistic regression.

\noindent\textbf{Knowledge Base (KB)}: KB is composed of a set of hard and soft rules modeling spatio-temporal interactions between various entities and the temporal structure of various complex events. The hard rules are assertions that should be strictly followed. Violation of hard rules sets the probability of the complex event to zero. On the other hand, soft rules allow uncertainty and exceptions. Violation of soft rules will make the complex event less probable but not impossible. 

\noindent\textbf{Markov Network (MN)}: Constants generated from visual processing step are used to instantiate (referred to as grounding) the variables in the KB rules. The grounded predicates for a Markov Random Field, referred to as Markov Network(MN). KB can be thought as template for constructing the Markov network. For every set of constants (detected visual entities and atomic events) observed in a scene, the FOPL rules involving the corresponding variables are instantiated to form the Markov network. Each node in MN represents either a grounded predicate or an inferred predicate. An edge exists between two nodes if the predicates appear in a formula. From the grounded network, MAP inference can be run to infer probabilities of query nodes after conditioning them with observed nodes and marginalizing out the hidden nodes.

As very few data sets offer complex events across a network of sensors, for training and evaluation we collected our own data from a network of four sensors with non-overlapping fields-of-view (see \ref{fig:MLNProcessing}{\it(Right)}). The data contained a variety of activities involving multiple human and vehicle agents. Targets detected from multiple sensors are associated across multiple sensors using appearance, shape and spatial-temporal cues. The Homography is estimated by manually labeling correspondence between the image and the ground map(done only once). The coordinated activities include :  dropping bag in a building and stealing bag from a building.
\Fig{fig:MLNProcessing} shows the entire processing pipeline. Specifically, we apply MLN to perform three key tasks: (a) Semantic scene labeling (see Section\ref{sec:sceneLabel}); (b) Target association across visual sensors (see Section\ref{sec:targetAssociation}) ; (c) Probabilistic fusion for detecting Complex events(see Section\ref{sec:complexEvents}).  Next section discusses the theoretical underpinnings of MLN.  

\section{Markov Logic Networks}
\label{sec:MLNTheory}
MLN allows multiple KB to be combined into a compact probabilistic model by assigning weights to the formulas, and is supported by a large range of learning and inference algorithms. Not only the weights but also the rules can be learned from the data set using Inductive logic programming(ILP). As the exact inference is intractable, Gibbs sampling (MCMC process) is used for performing the approximate inference. The rules in MLN form a template for constructing the Markov Network (MN) from the evidence. Evidence are in the form of grounded predicates, obtained by instantiating variables using all possible observed constants. The truth assignment for each of the predicates of the MRF defines a possible world $x$. The probability distribution over the possible worlds $\WW$, defined as joint distribution over the nodes of the corresponding MRF network, is the product of potentials associated with the cliques of the Markov Network:  
\begin{equation}
P(\WW = x) = \frac{1}{Z}\prod_k \phi_k(x_{\{k\}}) = \frac{1}{Z} \mbox{exp}\left(\sum_k w_k f_k(x_{\{k\}})\right)   
\end{equation}
where $x_{\{k\}}$ denotes the truth assignments of the nodes corresponding to $k^{th}$ clique of the MRF and $\phi_k(x_{\{k\}})$ is the potential function associated to the $k^{th}$ clique. Note that a clique in MRF corresponds to a grounded formula of the MLN. $f_k(x)$ is the feature associated to the $k^{th}$ clique and is 1 if the associated grounded formula is true and 0 otherwise, for each possible state of the nodes in the clique. The weights associated to the $k^{th}$ formula $w_k$, can be assigned manually or learned. This can be reformulated as:

\begin{tabular}{ll}
$P(\WW = x) $ & $=\frac{1}{Z}\mbox{exp}\left(\sum_k w_k f_k(x)\right)$ \\
   & $= \frac{1}{Z}\mbox{exp}\left(\sum_k w_k n_k(x)\right)$ \\ 
\end{tabular}

where $n_k(x)$ is the number of the times $k^{th}$ formula is true for different possible states of the nodes corresponding the $k^{th}$ clique $x_{\{j\}}$.  Z in the above equations refers to the partition function and is not used in the inference process, that involves maximizing the log-likelihood function. The equations simply represent that if the $k^{th}$ rule with weight $w_k$ is satisfied for a given set of constants and grounded atoms, the corresponding world  is $exp(w_k)$ times more probable than when the rule $k^{th}$ is not satisfied. For detecting occurrence of an activity, we query the MLN using the corresponding predicate. Given a set of evidence predicates $x=e$, hidden predicates $u$ and query predicates $y$, inference involves evaluating the MAP (Maximum-A-Posterior) distribution over query predicates $y$ conditioned on the evidence predicates $x$ and marginalizing out the hidden nodes $u$ as $P(y|x)$ :
\begin{equation}
\mbox{arg max}_y  \frac{1}{Z_x} \sum_{u \epsilon \{0,1\}}\mbox{exp}\left(\sum_k w_k n_k(y, u , x = e)\right)
\end{equation}

\noindent\textbf{Weights Learning in MLN}: MLN supports both generative and discriminative weights learning. Generative learning involves maximizing the log of the likelihood function to estimate the weights of the rules.
Unlike the inference process, that ignores the partition function Z, the gradient computation uses partition function Z. Even for reasonably sized domains, optimizing log-likelihood is intractable as it involves counting number of groundings $n_i(x)$ in which $i^{th}$ formula is true. 
Therefore, instead of optimizing likelihood, generative learning in existing implementation uses pseudo-log likelihood (PLL). Difference between PLL  and log-likelihood is that instead of using chain rule to factorize the joint distribution over entire nodes, we use Markov blanket to factorize the joint distribution into conditionals. The advantage of doing this is that predicates that do not appear in the same formula as a node can be ignored. This can speed up inference greatly. We use similar approach to scale inference to support multiple activities and longer videos. Discriminative learning on the other hand maximizes the conditional log-likelihood(CLL) of the queried atom given the observed atoms. The set of queried atoms need to be specified for discriminative learning. All the atoms are partitioned into observed $\XX$ and queried $\YY$. 

Discriminative learning maximizes the following conditional log-likelihood (CLL): 
\begin{equation}
P(Y = y | X = x) = \frac{1}{Z_x}\mbox{exp}\left(\sum_i w_i n_i(x,y)\right) 
\end{equation}			 			    
here $n_i(x,y)$ are the number of true groundings of the ith formula (composed of both queried and observed predicates). CLL is easier to optimize compared to the combined log-likelihood function of generative learning as the evidence constrains the probability of the query atoms to a much fewer possible states. Note that CLL and PLL optimization are equivalent when evidence predicates include the entire Markov Blanket of the query atoms.  A number of gradient•¡¹based optimization techniques exist(voted perceptron, contrastive divergence, diagonal Newton method and scaled conjugate gradient) for minimizing negative CLL. Singla and Domingos \cite{SD05} showed that learning weights by optimizing the CLL gives more accurate estimates of weights compared to PLL optimization. In this work we only used discriminative learning for estimating weights.
\begin{figure}[h]
\begin{center}
\includegraphics[width=0.95\linewidth]{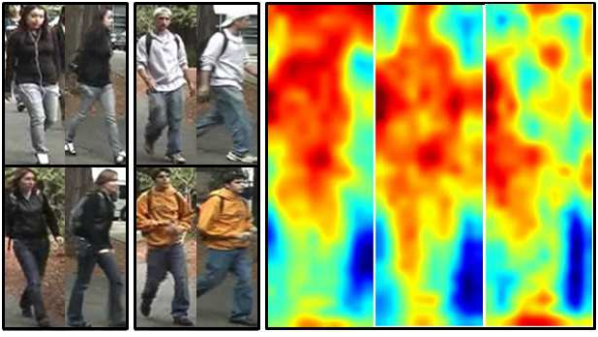}
\end{center}
\caption{{\it(Left)} Sample images from ViPER\cite{VIPER} dataset used as benchmark dataset  for evaluating appearance based matching of human targets; {\it(Right)} Top three relevance score masks for the patches obtained from RCA.}
\label{fig:RCAMask}
\end{figure}
\section{Scene Interpretation}
\label{sec:sceneLabel}
Contextual relations between the scene elements and the entities provide useful information about an activity occurring in a scene. We use domain knowledge, such as {\it humans can disappear only if they go out of scene or at an entrance of a building} and {\it human and vehicle footprints are more likely to be on a ground plane}, to formulate weighted rules in MLN and perform functional scene labeling of the image regions or refine scene element classification. Our scene analysis module first segments an image into multiple zones based on appearance cues using \cite{FH04}. We use Hoiem {\it et} al. \cite{HEH07} to categorize the image segments(see \fig{fig:BuildingEntryExit} top and middle rows) into one of the three categories $\cC = \{SKY, VERTICAL, HORIZONTAL\}$. Semantic scene labels can then be used to improve target association across sensors by enforcing spatial constraints on the targets such a human can only appear in image entry region. To that end, we automatically infer probability map of the entry or exit regions of the scene by formulating following rules: 
{
\begin{lstlisting}[
  mathescape,
  columns=fullflexible,
  basicstyle=\fontfamily{lmvtt}\selectfont,
] 
// Image regions where targets appear/dissapear 
// are entryExitZones(...)
$W_1$:  appearI(agent1,z1) $\rightarrow$ entryExitZone(z1)
$W_1$:  disappearI(agent1,z1) $\rightarrow$ entryExitZone(z1)

// Include adjacent regions with lower weights 
$W_2$:  appearI(agent1,z2) $\Lambda$ zoneAdjacentZone(z1,z2) 
		$\rightarrow$ entryExitZone(z1)
$W_2$:  disappearI(agent1,z2) $\Lambda$ zoneAdjacentZone(z1,z2) 
		$\rightarrow$ entryExitZone(z1)
\end{lstlisting}
}
here $W_2 < W_1$ to assign lower probability to the adjacent regions. Predicates $appearI(agent1,z1)$, $disappearI(agent1,z1)$ and $zoneAdjacentZone(z1,z2)$ are generated from the visual processing module, and represent if an agent appears or disappears in a zone, and whether two zones are adjacent to each other. The adjacency relation between a pair of zones, $zoneAdjacentZone(Z_1,Z_2)$, is computed based on whether the two segments lie near to each other (distance between the centroids) and if they share boundary. In addition to the spatio-temporal characteristics of the targets, scene elements classication scores are used to write more complex rules for extacting more meaningful information about the scene such as building entry/exit regions. Scene element classification scores can be easily ingested into the MLN inference system as soft evidences (weighted predicates) \textit{zoneClass(z, $\cC$)}. An image zone is a building(or garage) entry or exit region if it is a vertical structure and only human targets appear or disappear in those image regions. Additional probability may be associated to adjacent regions also :
{
\begin{lstlisting}[
  mathescape,
  columns=fullflexible,
  basicstyle=\fontfamily{lmvtt}\selectfont,
] 
// Regions  with human targets appear or disappear
zoneBuildingEntExit(z1) $\rightarrow$ zoneClass(z1,VERTICAL)
appearI(agent1,z1) $\Lambda$ class(agent1,HUMAN) 
		        $\rightarrow$ zoneBuildingEntExit(z1)
disappearI(agent1,z1) $\Lambda$ class(agent1,HUMAN) 
		 	$\rightarrow$ zoneBuildingEntExit(z1)

// Include adjacent regions also but with lower weights 
appearI(agent1,z2) $\Lambda$ class(agent1,HUMAN) $\Lambda$ 
  zoneAdjacentZone(z1,z2) $\Lambda$ zoneClass(z1,VERTICAL) 
			$\rightarrow$ zoneBuildingEntExit(z1)
disappearI(agent1,z2) $\Lambda$ class(agent1,HUMAN) $\Lambda$ 
  zoneAdjacentZone(z1,z2)$\Lambda$ zoneClass(z1,VERTICAL) 
			$\rightarrow$ zoneBuildingEntExit(z1)
\end{lstlisting}}
\Fig{fig:BuildingEntryExit}(Bottom row) shows the results of one of the camera image regions classified as building entry/exit, as obtained from MLN inference.
\section{Target Re-aquisition across Multiple Sensors}
\label{sec:targetAssociation} 
Targets detected in multiple sensors are fused in MLN using different entity similarity scores and spatial-temporal constraints, with the fusion parameters (weights) learned discriminatively using the MLN framework from a few labeled exemplars.
\subsection{Entity Similarity Relation Modeling}
Similarity relation modeling forms a critical component of information fusion systems in order to associate entities and events observed from data acquired from diverse and disparate sources.
Challenges to robust target similarity measure across different sensors include substantial variations resulting from the changes in sensor settings (white balance), illumination and viewing conditions, drastic changes in the pose and shape of the targets, and noise due to partial occlusions, cluttered backgrounds and presence of similar entities in the vicinity of the target. Invariance to some of these changes (such as illumination conditions) can be achieved using distance metric learning, that learns a transformation in the feature space such that image features corresponding to the same object are closer to each other.

\noindent\textbf{Entity Similarity Modeling using Metric Learning:} We employ metric learning approaches based on Relevance Component Analysis(RCA)\cite{HHSW05}, to enhance similarity relation between same entities when viewed under different imaging conditions. RCA seeks to identify and down-scale global unwanted variability within the data belonging to same class of objects.  The method transforms the feature space using a linear transformation by assigning large weights to the only relevant dimensions of the features and de-emphasizing those parts of the descriptor which are most influenced by the variability in the sensor data. 
For a set of N data points $\{(x_{ij}, j)\}$ belonging to K semantic classes with data points $n_j$, RCA first centers each data point belonging to a class to a common reference frame by subtracting in-class means $m_j$ (thus removing inter-class variability). It then reduces the intra-class variability by computing a whitening transformation of the in-class covariance matrix as: 
\begin{equation} 
C = \frac{1}{p} \sum_{(j=1)}^k \sum_{(i=1)}^{n_j}\left(x_{ji}- m_j\right)\left(x_{ji}- m_j\right)^t
\end{equation}
The whitening transform of the matrix, $W= C^{-1/2}$, is used as the linear transformation of the feature subspace such that features corresponding to same object are closer to each other. 
\begin{figure}[!ht]
\begin{center}
\includegraphics[width=0.99\linewidth]{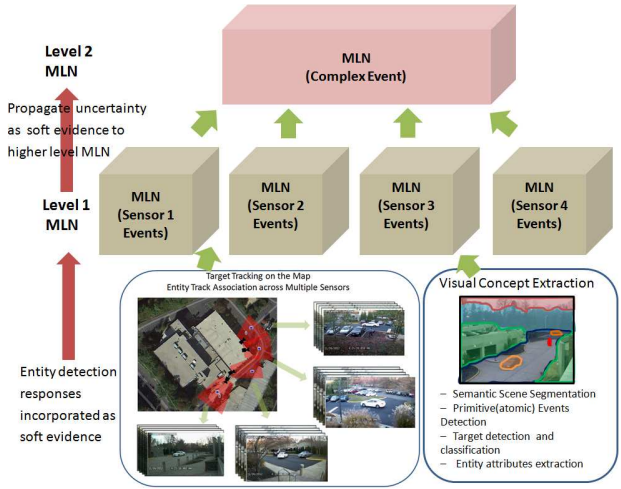}
\end{center}
\caption{Hierarchical splitting of a temporal sequence into multiple overlapping windows. Each of the box shown represents an MLN that fuses information from the MLNs from lower levels. Section \ref{subsec:events} discusses an example of a multi-level MLN for inferring a complex activity.}
\label{fig:HierarchicalMLN}
\end{figure}
\subsection {Target Association using MLN}
\label{subsec:TargetAssociation}
We apply MLN inference to associate the trajectories of the tracked targets across multiple cameras. Zhang et al.\cite{ZLN08} solved the global data association problem by formulating it as a min-cut/max-flow network. However, the method cannot handle long-term occlusions and provides only a single solution. Instead, we develop a solution based on MLN to perform data association and handle the problem of long-term occlusion across multiple sensors, while maintaining the multiple hypotheses for associations. The soft evidence of association is outputted as the predicate $equalAgent(...)$ with similarity score recalibrated to probability value, and used in high-level inference of activities. In the past, Leung and Herbin\cite{LH11} have employed MLN for data association of tracklets within a single sensor, to improve tracking performance. They adopt a simplistic approach of connecting two tracklets when their spatio-temporal coherence is less than a threshold value. However in our framework, we first learn the weights of MLN rules that govern the fusion of spatial, temporal and appearance similarity scores to determine equality of two entities observed in two different sensors. We use a subset of videos with labeled target associations to discriminatively train our MLN models.   

Tracklets extracted from Kalman filtering are used to perform target associations. Set of tracklets across multiple sensors are represented as $X={x_i}$, where a tracklet $x_i$ is defined as $x_i = f(c_i, t^{s}_i,  t^{e}_i, l_i, s_i, o_i, a_i)$. Here $c_i$ is the sensor ID, $t^{s}_i$ is the start time, $t^{e}_i$ is the end time,  $l_i$ is the location in the image or the map, $o_i$ is the class of the entity (human or vehicle), $s_i$ is the mensurated Euclidean 3D size of the entity (only used for vehicles)\cite{CRTH11}, and $a_i$ is appearance model of the target entity. The MLN rules for fusing multiple cues for the global data association problem are:
{
\begin{lstlisting}[
  mathescape,
  columns=fullflexible,
  basicstyle=\fontfamily{lmvtt}\selectfont,
] 
 $W_1$: temporallyClose($t^{e}_i$, $t^{s}_j$) $\rightarrow$ equalAgent($x_i$,$x_j$)
 $W_2$: spatiallyClose($l_i$, $l_j$) $\rightarrow$  equalAgent($x_i$,$x_j$)
 $W_3$: similarSize($s_i$, $s_j$)   $\rightarrow$  equalAgent($x_i$,$x_j$)
 $W_4$: similarClass($o_i$, $o_j$)  $\rightarrow$  equalAgent($x_i$,$x_j$)
 $W_5$: similarAppearance($o_i$, $o_j$)  $\rightarrow$  equalAgent($x_i$,$x_j$)
 $W_6$: temporallyClose($t^{e}_i$, $t^{s}_j$) $\Lambda$ spatiallyClose($l_i$, $l_j$) $\Lambda$ 
similarSize($s_i$, $s_j$)$\Lambda$ similarClass($o_i$, $o_j$) $\Lambda$ 
similarAppearance($o_i$, $o_j$) $\rightarrow$ equalAgent($x_i$,$x_j$)
\end{lstlisting}}
Note here that rules corresponding to individual cues have weights $\{W_i: i = 1,2,3,4,5\}$ that are usually lower than $W_6$ which is a much stronger rule and therefore carries larger weight. The rules yield a fusion framework that is somewhat similar to the posterior distribution defined in \cite{CRTH11}. However, here we are also learning the weights corresponding to each of the rules using only a few labeled examples. Next, we discuss the computation of the similarity predicates for each of the five similarity cues used for target association across sensors.
\begin{figure}
\begin{center}
\includegraphics[width=0.99\linewidth]{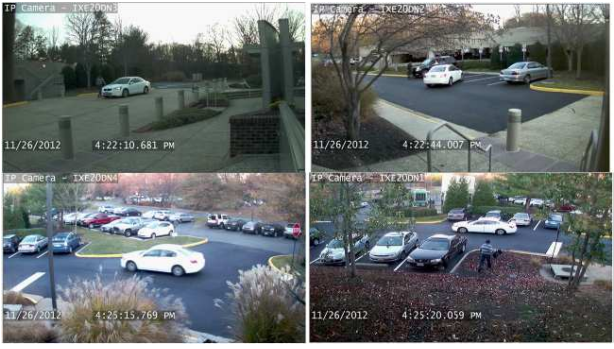}
\end{center}
\caption{Snapshots of video showing {\it bagStealEvent(...)}, acquired from 4 camera sensors with non-overlapping fields of view}
\label{fig:MultiCameraData}
\end{figure}

\noindent\textbf{Temporal Constraints:} This models temporal difference between the end and start time of a target across a pair of cameras using Gaussian distribution: 
\begin{equation}
temporallyClose(t^{A,e}_i, t^{B,s}_j) = \cN(f(t^{A,e}_i, t^{B,s}_j); m_t, \sigma^2_t)
\end{equation} 
For the non-overlapping sensors,  $f(t^{e}_i, t^{s}_j)$ computes this temporal difference.  If two cameras are nearby and there is no traffic signal between them, the variance tends to be smaller and contribute a lot to the similarity measurement. However, when two cameras are further away from each other or there are traffic signals in between, this similarity score will contribute less to the overall similarity measure since the distribution would be widely spread due to large variance.

\noindent\textbf{Spatial Constraints:} The spatial distance between objects in the two cameras is measured at the enter/exit regions of the scene. For a road with multiple lanes, each lane can be an enter/exit area. We apply MLN inference to directly classify image segments into enter/exit areas as discussed in section \ref{sec:sceneLabel}. The spatial probability is defined as: 
\begin{equation}
spatiallyClose(l^{A}_i , l^{B}_j) = \cN(dist(g(l^{A}_i), g(l^{B}_j)); m_l, \sigma^2_l)
\end{equation} 
Enter/exit areas of a scene are located mostly near the boundary of the image or at the entrance of a building. Function $g$ is the homography transform to project image locations $l^B$ and $l^A$ to map. Two targets detected in two cameras are only associated if they lie in the corresponding enter/exit areas.
 
\noindent\textbf{Size Similarity:} The size similarity score is computed for vehicle targets where we fit a 3D vehicle shape model \cite{CRTH11} to the silhouette of the target. The probability is computed as : 
\begin{equation}
similarSize(s^{A}_i , s^{B}_j) = \cN( \| s^{A}_i - s^{B}_j \|; m_s, \sigma^2_s )
\end{equation} 
3D vehicle fitting requires estimation of full projective transform which is performed manually by fitting a 3D vehicle model(sedan) with known dimensions to an image target. 

\noindent\textbf{Classification Similarity:} For computing classification similarity $similarClass(o^A_{j} , o^B_{j})$, we first characterize the empirical probability of classifying a target for each of the visual sensor, as classification accuracy depends on the camera intrinsics and calibration accuracy. Empirical probability is computed from the class confusion matrix for each sensor A where each matrix element $\cC^A_{i,j}$ represents probability $P(o^A_{j} | c_{i})$ of classifying object $j$  to  class $i$. For computing the classification similarity we assign higher weight to the camera with higher classification accuracy. The joint classification probability of the same object observed from camera $A$ and $B$ is defined as:
\begin{equation}
 P(o^A_j , o^B_j) = \sum\limits_{k=N} P(o^A_{j} , o^B_{j} | c_k) P(c_k)
\end{equation} 
\indent where $o^A_{j}$ and  $o^A_{j}$ are the observed classes and $c_k$ is the groundtruth. Classification in each sensor is conditionally independent given the object class, the similarity measure can be computed as:

\begin{equation} 
P(o^A_j , o^B_j) = \sum\limits_{k=N} P(o^A_{j} | c_k) P(o^B_{j} | c_k) P(c_k)
\end{equation}
\indent where $P(o^A_{j} | c_k)$ and $P(o^B_{j} | c_k)$ can be computed from the confusion matrix, and $P(c_k)$ can be either set to uniform or estimated as the marginal probability from the confusion matrix.  

\noindent\textbf{Appearance Similarity for Vehicles and Humans:} Since vehicles exhibit significant variation in shapes due to viewpoint changes, shape based descriptors did not improve matching scores. Covariance descriptor\cite{TPM06} based on only color, gave sufficiently accurate matching results for vehicles across sensors. Humans exhibit significant variation in appearance compared to vehicles and often have noisier localization due to moving too close to each other, carrying an accessory and forming significantly large shadows on the ground. For matching humans however, unique compositional parts provide strongly discriminative cues for matching, and has already been applied in some of the recent works\cite{ZOW13}. Our algorithm extends this work and computes similarity scores between target images by matching densely sampled patches within a constrained search neighborhood (longer horizontally and shorter vertically). The matching score is boosted by the saliency score $\cS$ that characterizes how discriminative a patch is based on its similarity to other reference patches. A patch exhibiting larger variance for the K nearest neighbor reference patches is given higher saliency score $\cS(\xx)$. In addition to the saliency, in our similarity score we also factor in a relevance based weighting scheme to down weigh patches, that are predominantly due to background clutter. We use RCA to obtain such a relevance score $\cR(\xx)$ from a set of training examples. The similarity measure between the two images, $\xx^p$ and $\xx^q$, is computed as : $Sim(\xx^p,\xx^q)=$ 
\begin{equation}
\sum_{m,n} \frac{\cS(x^p_{m,n})\cR(x^p_{m,n})d(x^p_{m,n},x^q_{m,n}))\cS(x^q_{m,n})\cR(x^q_{m,n})}{\al + |\cS(x^p_{m,n}) - \cS(x^q_{m,n})| }
\end{equation} 
where  $x^p_{m,n}$ denote $(m,n)$ patch from the image $p$, $\al$ is the normalization constant, and the denominator term penalizes large difference in saliency scores of two patches. RCA uses only positive similarity constraints to learn a global metric space such that intra-class variability is minimized. Patches corresponding to highest variability are due to the background clutter and are automatically down weighed during matching. The relevance score for a patch is computed as absolute sum of vector coefficients corresponding to that patch for the first column vector of the transformation matrix. \Fig{fig:RCAMask} shows the relevance mask learned using RCA for the Viper data set used for benchmarking proposed algorithm with the \cite{ZOW13}. Appearance similarity between targets are used to generate soft evidence predicates $similarAppearance(a^A_i, a^B_j)$ for associating target $i$ in camera A to target $j$ in camera B.   
\begin{table}[t] 
{\small
\begin{center}
\begin{tabular}{|l|l|}
\hline
\textbf{Event Predicate} & \textbf{Description about the Event}\\
\hline 
\textit{zoneBuildingEntExit(Z)}  & Zone is a building entry exit \\ 
\textit{zoneAdjacentZone($Z_1$,$Z_2$)} & Two zones adjacent to each other\\
\hline 
\textit{humanEntBuilding(..)} & Human enters building \\ 
\textit{parkVehicle(A)}  & Vehicle arriving in the parking lot \& \\ 
                         & stopping in the next time interval \\ 
\textit{driveVehicleAway(A)} & Stationary vehicle that starts \\
			     & moving in the next time interval \\
\textit{passVehicle(A)}  & Vehicle observed passing across  \\ 
			 & camera                      \\
\hline 
\textit{embark(A,B)}     & Human A comes near vehicle B and \\ 
                         & disappears after which vehicle B \\
			 & starts moving \\
\textit{disembark(A,B)}  & Human target appears close to a \\
                         & stationary vehicle target\\
\textit{embarkWithBag(A,B)}  & Human A with {\it carryBag(...)}\\ 
                         & predicate embarks a vehicle B \\
\hline
\textit{equalAgents(A,B)} & Agents A and B  across different\\ 
                          & sensors are same(Target association) \\
\hline 
\textit{sensorXEvents(...)} & Events observed in sensor X\\
\hline
\end{tabular}
\end{center}}
\caption{Unobserved predicates representing various sub-events that are used as inputs for high-level analysis and detecting a complex event across multiple sensors}
\label{tab:predicates}
\end{table}

\section{Formulation of MLN Rules}
\label{sec:complexEvents}
MLN allows principled data fusion from multiple sensors, while taking into account the errors and uncertainties, and achieving potentially more accurate inference over doing the same using individual sensors. The information extracted from different sensors differs in the representation and the encoded semantics, and therefore should be fused at multiple levels of granularity. Low level information fusion would combine primitive events, local entity interactions in a sensor to infer sub-events. Higher level inference for detecting complex events will progressively use more meaningful information as generated from low-level inference to make decisions. Uncertainties may introduces at any stage due to missed or false detection of targets and atomic events, target tracking and association across cameras and target attribute extraction. To this end, unlike past MLN based activity recognition frameworks\cite{TD08,KYD10,MD11}) that used hard evidences, we generate predicates with an associated probability(soft evidence). The soft evidence thus enables propagation of uncertainty from the lowest level of visual processing to high-level decision making.  
\begin{figure}[h]
\begin{center}
\includegraphics[width=0.99\linewidth]{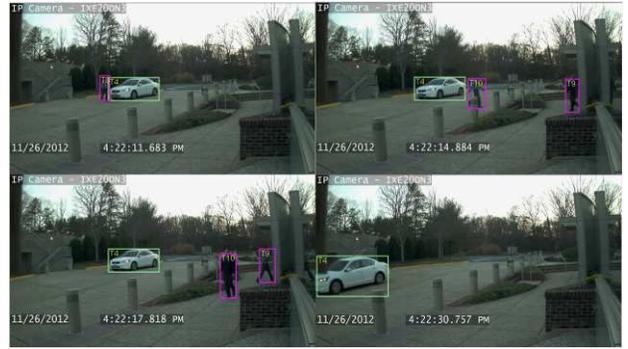}
\end{center}
\caption{Example of subevents detected in sensor 1 for the complex event $bagStealEvent(\cdots)$. The subevents are $parkVehicle(..)$ , $disembark(..)$ , $humanEntBuilding(..)$ and $driveVehicleAway(..)$}
\label{fig:sensor1EventsResult}
\end{figure}

\subsection{Events Modeling and Recognition}
\label{subsec:events}
The visual processing module generates groundings at fixed time intervals by detecting and tracking the targets in the videos. The generated constants include - sensor ids, agent ids, zones ids and types (for semantic scene labeling tasks), agent class types, location and time. Spatial location is a constant pair $Loc\_X\_Y$ either as an image pixel coordinates or on the ground map obtained using image to map homography. The time is represented as an instant or as an interval using starting and ending time $TimeInt\_S\_E$. We detect two classes of agents in the scene - vehicles and humans. Image zones are categorized into one of the three geometric classes {\it \cC} classes. The grounded atoms are intantiated predicates and represent either an agent attribute or any primitive event it is performing. The ground predicates include: (a) zone classifications $zoneClass(Z_1,ZType)$ ; (b) zone where an agent appears $appearI(A_1,Z_1)$ or disappears $disappearI(A_1,Z_1)$ ; (c)  agent classification $class(A_1,AType)$ ; (d) primitive events $appear(A_1$ , $Loc,Time)$, $disappear(A_1$ , $Loc, Time)$ , $move(A_1, LocS$ , $LocE,TimeInt)$ and $stationary(A_1$ , $Loc,TimeInt)$ ; and (e) agent is carrying a bag $carryBag(A_1)$. The grounded predicates and constants generated from the visual processing module are used to generate Markov Network. 
\begin{figure}[!ht]
\begin{center}
\includegraphics[width=1\linewidth]{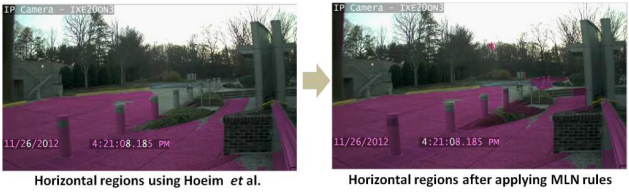}
\end{center}
\caption{Effect of applying domain knowledge to refine geometric labels {\it (HORIZONTAL)} of the scene. We use presence of human and vehicle target footprints to improve classification confidence scores of {\it HORIZONTAL} class}
\label{fig:GroundPlaneRefinement}
\end{figure}
Complex events are recognized by querying for the corresponding unobserved predicates, running the inference using fast Gibbs sampler and estimating their probabilities. These predicates involve both unknown hidden predicates that are marginalized out during inference and the queried predicates. We list the relevant predicates along with their description in the table \ref{tab:predicates}. We applied MLN inference to detect two different complex activities that are composed of sub-events listed in table \ref{tab:predicates}: 
\begin{enumerate}
\item {\it bagStealEvent(...)}: Vehicle appears in sensor $C_1$, a human disembarks the vehicle and enters a building. Vehicle drives away and parks in sensor C2 field of view. After sometime vehicle drives away and is seen passing across sensor $C_3$. It appears in sensor $C_4$ where the human reappears with a bag and embarks the vehicle. The vehicle drives away from  sensor.   
\item {\it bagDropEvent(...)}: The sequence of events are similar to {\it bagStealEvent(...)} with the difference that human enters the building with a bag in sensor $C_1$ and reappears in sensor $C_2$ without a bag.
\end{enumerate} 

Complex activities are spread across network of four sensors and involve interactions between multiple agents, a bag and the environment. For each of the activities, we first identify a set of sub-events that are detected in each sensor (denoted by {\it sensorXEvents(...)}). The MLN rules for detecting sub-events for the complex event {\it bagStealEvent(...)} in sensor $C_1$ are 
{\small
\begin{lstlisting}[
  mathescape,
  columns=fullflexible,
  basicstyle=\fontfamily{lmvtt}\selectfont,
] 
 disembark($A_1$,$A_2$,$Int_1$,$T_1$) $\Lambda$ humanEntBuilding($A_3$,$T_2$) $\Lambda$ 
 equalAgents($A_1$,$A_3$) $\Lambda$ driveVehicleAway($A_2$,$Int_2$) $\Lambda$ 
 sensorType($C_1$) $\rightarrow$ sensor1Events($A_1$,$A_2$,$Int_2$)
\end{lstlisting}}
The predicate {\it sensorType(...)} is to enforce hard constraints that only constants generated from sensor $C_1$ are used for inference of the query predicate. Each of the sub-events are detected using MLN inference engine associated to each sensor(see \fig{fig:HierarchicalMLN}) and the result predicates are fed into higher level MLN along with the associated probabilites, for inferring complex event. The rule formulation of the {\it bagStealEvent(...)} activity are as follows:
{\small
\begin{lstlisting}[
  mathescape,
  columns=fullflexible,
  basicstyle=\fontfamily{lmvtt}\selectfont,
] 
sensor1Events($A_1$,$A_2$,$Int_1$) $\Lambda$ sensor2Events($A_3$,$A_4$,$Int_2$) $\Lambda$ 
afterInt($Int_1$,$Int_2$) $\Lambda$ equalAgents($A_1$,$A_3$) $\Lambda$ $\cdots$ $\Lambda$
sensorNEvents($A_M$,$A_N$,$Int_K$) $\Lambda$ afterInt($Int_K$-1,$Int_K$) $\Lambda$ 
equalAgents($A_M$-1,$A_M$) $\rightarrow$ ComplexEvent($A_1$,$\cdots$,$A_M$,$Int_K$)
\end{lstlisting}}
First order predicate logic (FOPL) rule for detecting generic complex event involving multiple agents and target association across multiple sensors. For each sensor, we define a predicate for events occurring in that sensor. The agents in that sensor are associated to the other sensor using target association MLN engine (that infers {\it equalAgent(...)} predicate). The predicate {\it afterInt($Int_1$,$Int_2$)} is true if the time interval {\it $Int_1$} occurs before the {\it $Int_2$}.

\subsection{Complexity of Inference using MLN}
Inference in MLN is $\#$P- hard problem, with no polynomial time algorithm for exactly counting the number of true cliques(representing instantiated formulas) in the grounded network(an MRF). The nodes in MN grows exponentially with the number of instances and formulas in the Knowledge Base(KB). Since all the constants are used to instantiate all the variables of the same type, in all the predicates used in the rules, predicates with high arity cause combinatorial explosion in the number of possible cliques formed after the grounding step. Similarly long rules also cause high order dependencies in the relations and larger cliques in MN. No exact algorithm is available for minimization of the energy function involving high-order cliques.
\begin{figure}[h]
\begin{center}
\includegraphics[width=0.95\linewidth]{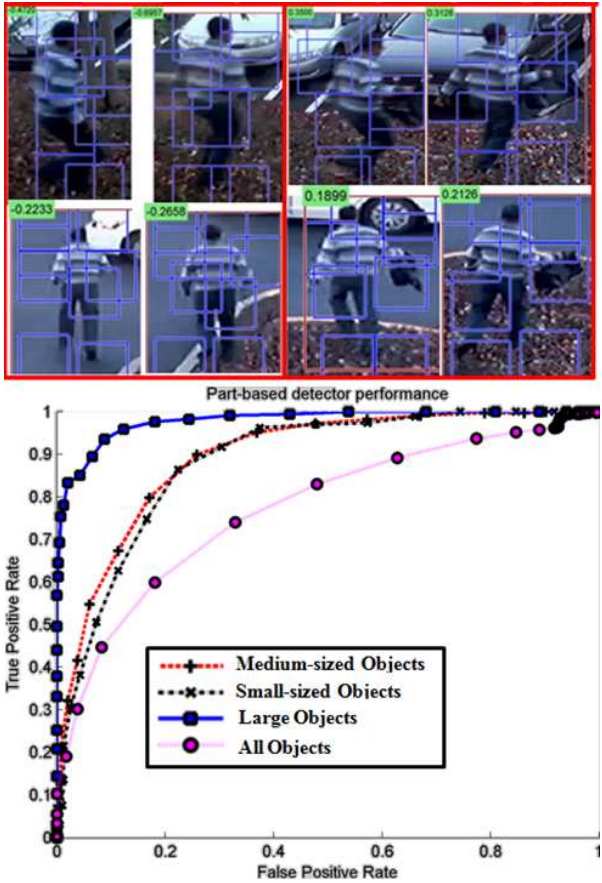}
\end{center}
\caption{
{\it (Top)} Top row shows the classifier responses on the two sequences containing human with and without bag from our dataset; {\it(Bottom)} Precision-Recall curve for bag detection iLIDS and Virat dataset for 3 class of bag sizes ;}
\label{fig:BagDetection}
\end{figure}

\noindent\textbf{MLN Implementation:} Alchemy\cite{DL09,TD08,MD11,KYD10} is an open-source system that provides a number of algorithms for statistical relational learning and inference based on the Markov logic representation. However Alchemy implementation suffers from many drawbacks: (a) top-down grounding leading to rapid increase in memory requirements even for small set of rules ; (b) no support for soft evidences in the inference and ; (c) non-scalable due to the requirement of having entire grounded network to be in the memory during inference. A more recent implementation of MLN, TUFFY\cite{NRDS11} overcomes all these limitations by providing bottom-up grounding by employing Relation Database Management System (RDBMS) as a backend tool for storage and query. The rules in the MLN are written to minimize combinatorial explosion during inference. TUFFY allows use of conditions, as the last component of either the antecedent or the consequent, to restrict the range of constants used for grounding a formula. Using hard constraints further also improves tractability of inference as an interpretation of the world violating a hard constraint has zero probability and can be readily eliminated during bottom-up grounding. Using multiple smaller rules instead of one long rule also improves the grounding by forming smaller cliques in the network and fewer nodes. We further reduce the arity of the predicates by combining multiple dimensions of the spatial location (X-Y coordinates) and time interval (start and end time) into one unit. This greatly improves the grounding and inference step. For example, the arity of the predicate $move(A$, $LocX1$, $LocY1$, $Time1$, $LocX2$, $LocY2$, $Time2)$ gets reduced to $move(A$,$LocX1\_Y1$, $LocX2\_Y2, IntTime1\_Time2)$.  

\noindent\textbf{Scalable Hierarchical Inference in MLN:} Inference in MLN for sensor activities can be significantly improved if instead of generating a single Markov Network(MN) for all the activities, we explicitly partition the MN into multiple activity specific networks containing only the predicate nodes that appear in only the formulas of the activity. This restriction effectively considers only the Markov Blanket(MB) of a predicate node for computing expected number of true groundings and had been widely used as an alternative to exact computation. From implementation perspective this is equivalent to having a separate MLN inference engine for each activities, and employing a hierarchical inference (see \fig{fig:HierarchicalMLN}) where the semantic information extracted at each level of abstraction is propagated from the lowest visual processing level to sub-event detection MLN engine, and finally to the high-level complex event processing module. Moreover, since the primitive events and various sub-events (as listed in Table \ref{tab:predicates}) are dependent only on temporally local interactions between the agents, for analyzing long videos we divide a long temporal sequence into multiple overlapping smaller sequences, and run MLN engine within each of these sequences independently. Finally, the query result predicates from each temporal windows are merged using a high level MLN engine for inferring long-term events extending across multiple such windows.  
A significant advantage of TUFFY is that it supports soft evidences that allows propagating uncertainties in the spatial and temporal fusion process used in our framework. Result predicates from low-level MLN are incorporated as rules with the weights computed as log odds of the predicate probability $\mbox{ln} \frac{p}{1-p}$. This allows partitioning the grounding and inference in MN in order to scale it to larger problems.   
\section{Experiments}
MLN inference engine was used for detecting complex events in a 1 hour long videos, acquired from 4 camera sensors (see \fig{fig:MultiCameraData}). The inference engine was run on windows of 15 mins each to maintain tractability of MLN inference. Each of the captured sequence were high-definition videos of resolution $1920 \times 1040$ and provided sufficient number of pixels to detect carrying bag attributes. 

\noindent\textbf{Semantic Scene Labeling:} For scene labeling we only show qualitative results. We ran image segmentation\cite{FH04} to generate a total of 217 segments(zones) in the image. Each segment is classified into various geometric classes as in $\cC$ using appearance and geometric cues \cite{HEH07} as discussed in section \ref{sec:sceneLabel}. As another example of our MLN based framework, we wrote MLN rules to incorporate the footprints of the targets in the image regions as additional cues to label them as {\it HORIZONTAL}. \Fig{fig:GroundPlaneRefinement} shows the effect of applying this domain knowledge to refine labels of some difficult to classify regions.
\begin{figure*}[!ht]
\begin{center}
\includegraphics[width=0.9\linewidth]{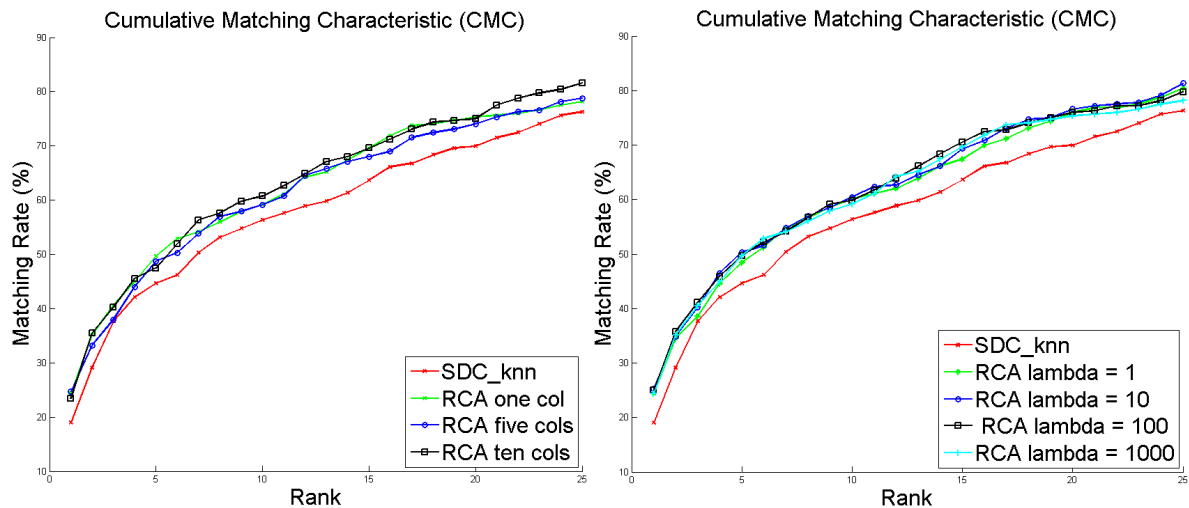}
\end{center}
\caption{Effect of RCA on the cumulative matching performance of SDC\cite{ZOW13}. {\it(left)} Effect of RCA when 1, 5 and 10 columns of the transformation learned from RCA are used for computing the patch relevance score ; {\it(right)} Effect of varying the regularization parameter for RCA computation}
\label{fig:RCASDC}
\end{figure*}

\noindent\textbf{Entity Attribute Extraction:} Our system currently only detects whether a human target is carrying bag or not. We trained Deformable Part-based Model (DPM)\cite{FMR08} based classifiers for detecting humans carrying bag. Our training dataset contained  624 positive exemplars and  265 negative exemplars. For testing we used 502 positive exemplars and 571 negative exemplars. We further subcategorized bags into large(being pushed or dragged), medium(carry bags) and small (held along). We evaluated the performance of the attribute detector on Virat  dataset\cite{VIRAT} and iLIDS\cite{ILIDS}. \Fig{fig:BagDetection}(left) shows the true positive rates (TPR) and false positive rates (FPR) for the three classes of objects. On the right we show samples from the training dataset and classification response on our complex activity sequence. 
\begin{figure}
\begin{center}
\includegraphics[width=0.95\linewidth]{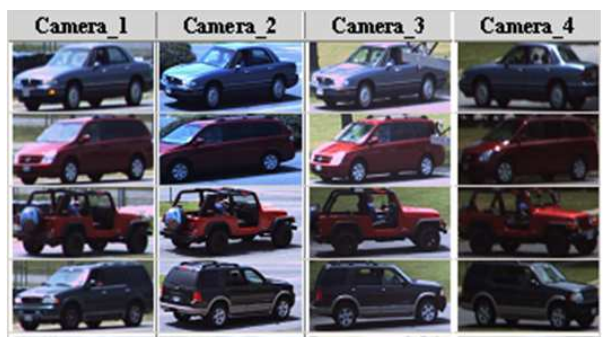}
\end{center}
\caption{The vehicle images acquired from the 4 sensors of our \textit{MultiCamera} dataset with non-overlapping fields of view}
\label{fig:MultiCamera}
\end{figure}

\noindent\textbf{Vehicle Track Association:} As our current dataset did not have significant number of vehicular targets, we evaluated our vehicle association algorithm on a separate \textit{MultiCamera} dataset of 24-minute durations, $4000\times640$ pixel resolution, taken from four different locations with non overlapping fields of view, and contained 154 vehicle targets. \Fig{fig:MultiCamera} shows the snapshots of the same target observed from the 4 sensors. This dataset is challenging due to: i) cameras being more that 750 meters apart and have traffic signals in between ; ii) less reliable classification due to inaccurate projection matrix for each camera ; iii)large variation in color and illumination changes; iv) not all vehicles passed through all the 4 cameras.
Table \ref{tab:DDatasetResults} shows the effect of applying distance metric learning in improving the precision-recall rates of targets correctly associated across cameras. Our similarity measure is composed of appearance as well as class, size and spatial-temporal constraints (see section \ref{subsec:TargetAssociation}). For computing appearance of vehicles we evaluated covariance descriptor of color histogram and gradient filter responses, with and without distance metric learning(RCA). We used 60 bases vectors learned using RCA for transforming the color histogram space and 600 dimensions for the covariance descriptor computed from image gradients. The results of tracking accuracy are shown in table \ref{tab:DDatasetResults}. We also show the accuracy using F-measure (or F-score), defined as $\frac{2(Recall \times Precision)}{(Recall + Precision)}$. The results show that use of RCA improves the overall precision/recall, and gradient information only marginally improved the precision while maintaining the same level of recall rates. The overall improvement in the F-measure after applying RCA was from 79.57\% to 84.26\%. Also note that including an appearance model without RCA did not improve F-score significantly.
\begin{table}
\begin{center}
{\small
\begin{tabular}{ | c | c | c | c | c |}
\hline
Appearance     & No Appear- & Color     & RCA Only  & RCA Color \\ 
Model          & ance       & Hist.     & Color	    & +Gradient    \\ 
\hline
Recall	       & 0.778      & 0.778     & \textbf{0.835} & 0.815       \\ 
\hline
Precision      & 0.797	    & 0.815     & 0.846     & \textbf{0.872} \\ 
\hline
F-Measure      & 0.788	    & 0.796     & 0.841     & \textbf{0.843} \\
\hline
\end{tabular}}
\end{center}
\caption{Recall, precision and F-measure of multi-camera target tracking}
\label{tab:DDatasetResults}
\end{table}

\noindent\textbf{Human Track Association:} We compared the accuracy of human target association with the state-of-the-art Saliency Detection(SDC) based algorithm \cite{ZOW13} on the standard dataset ViPER\cite{VIPER}. The data contains 637 targets viewed from two different cameras from different viewpoints and at different time instants. We use Cumulative Matching Characteristic(CMC) metric (matching rate w.r.t. rank) to demonstrate the improvement in the accuracy. \Fig{fig:RCASDC} shows the plots obtained after using their similarity score and the plots after incorporating distance metric learning using RCA in the score. Note here that the results shown here did not use individual human masks used for generating results in \cite{ZOW13}. We show results for the cases when we compute relevance score using different number of columns (1,5 and 10), and also when we compute RCA transformation for different regularization parameters ``lambda''. Notice that RCA significantly boosts accuracy of the original algorithm\cite{ZOW13}.

\noindent\textbf{Complex Event Detection:} We evaluate our MLN based system for recognizing complex events on multi-sensor video sequence acquired from four sensors. Table \ref{tab:results} summarizes the complex event detection results of our system. We applied  MLN inference to detect 2 complex activities of \newline $bagStealEvent(\cdots)$ and $bagDropEvent(\cdots)$ involving multiple interacting targets and spread across 4 sensors. In order to avoid high computational cost of MLN inference due to large number of predicates, we adopt hierarchical processing (discussed in section \ref{subsec:events}) to perform inference in MLN. We used TUFFY\cite{NRDS11} for our MLN implementation. Although past works such as \cite{MD11} adopt similar approach for inference, our system employs soft evidences in the form of weighted predicates, that allowed principled propagation of uncertainty from the low-level of MLN inference for scene label refinement and visual processing, to the high-level MLN inference for inferring complex events. Inference of the long rule for $bagStealEvent(\cdots)$ typically takes 1 hour on 2.8 GHz quad core machine. Due to high arity of the predicate for $bagStealEvent(\cdots)$, it generates $\thicksim 400K$ possible ground atoms all possible combinations of constants in grounded atoms. Recognizing these complex events require detection of multiple sub-events as listed in the table \ref{tab:results}. Overall events detection precision was $77.8\%$ and the recall was $80.7\%$. The videos only contained 4 complex events 3 of which was successfully detected. The fourth activity of $bagDropEvent(..)$ was falsely detected as $bagStealEvent(..)$ due to inaccurate carrying bag attribute detector. Finally, in most of the events, recall was high while there were large number of false detections. This was primarily due to number of false human targets interacting with vehicle or building entrances. Inference in MLN is always performed on interval instances it has already seen in the visual predicates. MLN inference on a domain of time intervals will be intractable and therefore we report accuracy in terms of \# of detections only (and not on time intervals of the events). The weights of our soft rules were manually set while the weights of the MLN rules for target track association across different sensors were learned.   
\begin{table} 
{\scriptsize
\begin{center}
\begin{tabular}{|l|c|c|c|c|c|}
\hline
Events                 &  No. of  & True     & False     &  Prec.  &  Recall \\
                       &  Events  & Detect.  & Detect.   &         &         \\ 
\hline\hline
Park Vehicle           & 14       & 14      & 2          &  0.87   & 1.0    \\ 
Drive Vehicle          & 10       & 10      & 2          &  0.83   & 1.0    \\
Passing Vehicle        & 52       & 39      & 6          &  0.85   & 0.75   \\
\hline
Embarking              & 4        & 4        & 3         &  0.57   & 1.0    \\
Disembarking           & 5        & 5        & 3         &  0.63   & 1.0    \\
Enter Building         & 14       & 10       & 4         &  0.71   & 0.71   \\
Exit Building          & 10       & 6        & 5         &  0.54   & 0.6    \\
\hline
\# Sub-Events          & 109      & 88       & 25        & 0.78    &  0.81   \\
\hline
Steal Bag              & 2        & 2        & 1         &  0.66   & 1.0     \\
Drop Bag               & 2        & 1        & 0         &  1.0    & 0.5     \\
\hline
\end{tabular}
\end{center}}
\caption{Results for complex event recognition using MLN}
\label{tab:results}
\end{table}

\section{Conclusion} 
In this work we demonstrate a Markov Logic based decision system for recognizing complex events in videos acquired from a network of sensors. We apply Markov Logic Networks as a framework for representing and applying domain knowledge, a probabilistic fusion engine to combine information of varied modalities, and a tool for making decisions under uncertainty. We further enhance algorithms for modeling similarity relation between observations of an entity from different sensors. We also applied MLN to the problem of semantic and functional labeling of image regions in a scene. Finally, we demonstrated that decision systems based on Markov Logic Networks can be scaled to detect complex multi-agent activities in long video sequences using hierarchical inference without sacrificing prediction accuracy. 

\noindent\textbf{Acknowledgements:} This research was sponsored by Office of Naval Research under contract number N00014-11-M-0037.
{\bibliographystyle{abbrv}
\bibliography{PapBib}  
%
%
}

\end{document}